\documentclass[lettersize,journal]{IEEEtran}
\usepackage{amsmath,amsfonts}
\usepackage{algorithm}
\usepackage{algpseudocode}
\usepackage{array}
\usepackage[caption=false,font=normalsize,labelfont=sf,textfont=sf]{subfig}
\usepackage{textcomp}
\usepackage{stfloats}
\usepackage{url}
\usepackage{verbatim}
\usepackage{graphicx}
\usepackage{cite}

\usepackage[english]{babel}
\usepackage{booktabs} %
\usepackage{rotating} %
\usepackage{multirow} %
\usepackage{colortbl} %
\usepackage{xcolor} %
\usepackage{color}    %

\usepackage{amsmath}  %
\usepackage{amsfonts} %
\usepackage{amssymb} %
\usepackage{setspace}
\usepackage{adjustbox}
\usepackage{pifont} %
\usepackage{enumitem} %
\usepackage{array}

\usepackage{listings}

\newcommand{\noopsort}[1]{}

\usepackage{float}
\usepackage{hyperref} 
\usepackage{threeparttable}

\usepackage{makecell} %

\newcommand{\loldu}{LoLDU }
\newcommand{\loldunospace}{LoLDU}

\begin{document}

\title{\textbf{LoLDU}: \textbf{Lo}w-Rank Adaptation via \textbf{L}ower-\textbf{D}iag-\textbf{U}pper Decomposition for Parameter-Efficient Fine-Tuning}

\author{Yiming Shi, Jiwei Wei, Yujia Wu, Ran Ran, Chengwei Sun, Shiyuan He, Yang Yang

\IEEEcompsocitemizethanks{
	\IEEEcompsocthanksitem Yiming Shi, Jiwei Wei, Yujia Wu, Ran Ran, Chengwei Sun, Shiyuan He and Yang Yang are with the Center for Future Media and School of Computer Science and Engineering, University of Electronic Science and Technology of China, Chengdu 611731, China (e-mail: yimingshi666@gmail.com; mathematic6@gmail.com; 202322080314@std.uestc.edu.cn; ranran@std.uestc.edu.cn; suncw10@126.com).
	\IEEEcompsocthanksitem Corresponding author: Jiwei Wei. Email: mathematic6@gmail.com.
}}

\markboth{Journal of \LaTeX\ Class Files,~Vol.~14, No.~8, August~2021}%
{Shell \MakeLowercase{\textit{et al.}}: A Sample Article Using IEEEtran.cls for IEEE Journals}

\maketitle

\begin{abstract}
    The rapid growth of model scale has necessitated substantial computational resources for fine-tuning. 
    Existing approach such as Low-Rank Adaptation (LoRA) has sought to address the problem of handling the large updated parameters in full fine-tuning. However, LoRA utilize random initialization and optimization of low-rank matrices to approximate updated weights, which can result in suboptimal convergence and an accuracy gap compared to full fine-tuning.
    To address these issues, we propose LoLDU, a Parameter-Efficient Fine-Tuning (PEFT) approach that significantly reduces trainable parameters by 2600 times compared to regular PEFT methods while maintaining comparable performance. LoLDU leverages Lower-Diag-Upper Decomposition (LDU) to initialize low-rank matrices for faster convergence and orthogonality. We focus on optimizing the diagonal matrix for scaling transformations.
    To the best of our knowledge, LoLDU has the fewest parameters among all PEFT approaches. We conducted extensive experiments across 4 instruction-following datasets, 6 natural language understanding (NLU) datasets, 8 image classification datasets, and image generation datasets with multiple model types (LLaMA2, RoBERTa, ViT, and Stable Diffusion), providing a comprehensive and detailed analysis. Our open-source code can be accessed at \href{https://github.com/SKDDJ/LoLDU}{https://github.com/SKDDJ/LoLDU}. 
\end{abstract}


\section{Introduction}
\label{sec:introduction}

\IEEEPARstart{W}{ITHIN} the era of exponentially increasing the scale of models, fine-tuning these large models for new domains (e.g., Visual Instruction Tuning\cite{liuVisualInstructionTuning2023}),  applying advanced learning techniques (e.g., Representation Learning\cite{8945165, 8890009, 8886728}), or adapting to downstream tasks (e.g., Text-to-Image Customization\cite{gal2022imageworthwordpersonalizing,ruiz2023dreamboothfinetuningtexttoimage}, Object Tracking \cite{lai2024refocus, 9932644}) requires substantial computational resources. To address this challenge, Parameter-Efficient Fine-Tuning (PEFT) techniques such as LoRA\cite{huLoRALowRankAdaptation2021}, VeRA\cite{kopiczko2024vera},QLoRA\cite{dettmers2023qloraefficientfinetuningquantized}, and PiSSA\cite{mengPiSSAPrincipalSingular2024} have been developed to mitigate the bottleneck by reducing the number of trainable parameters, memory (VRAM), and storage costs.

\begin{figure}[t!]
    \centering
    \includegraphics[width=\columnwidth]{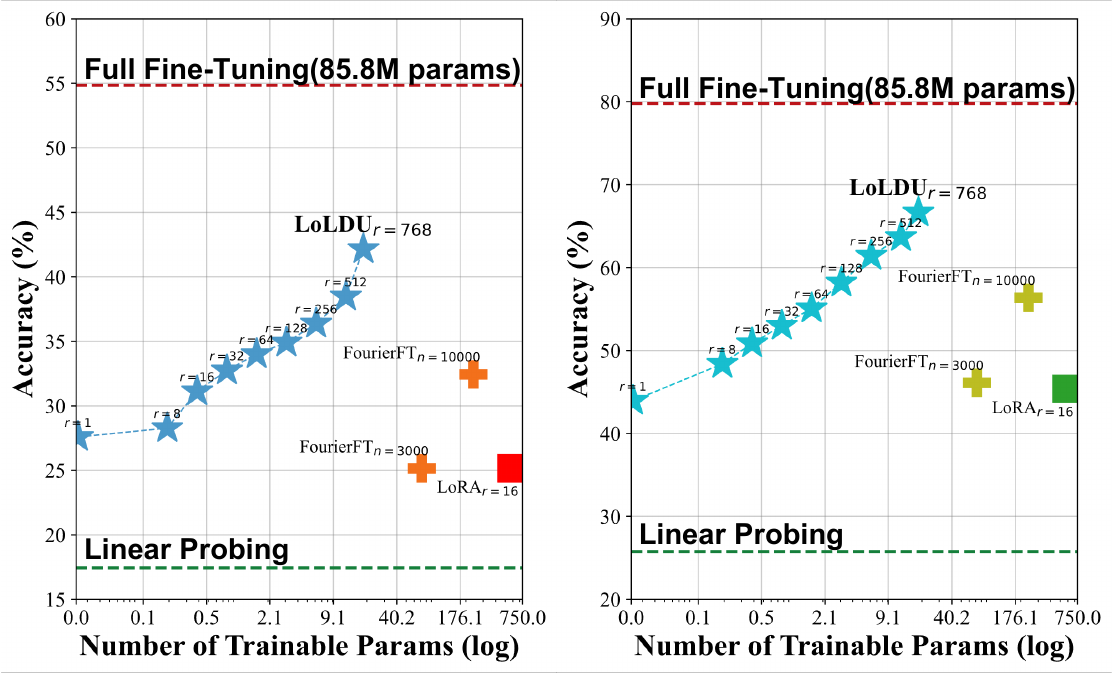}
    \vspace{-2em}
    \setcounter{footnote}{-1}
    \caption{\textbf{Performance vs log-scaled trainable parameters for FGVC (left) and StanfordCars (right) on ViT Base}. Our \loldu methods with $r=\{1,8,16,32,64,128,256,512,768\}$ exhibit superior parameter efficiency and performance when contrasted with  Linear Probing\cite{chen2021empiricalstudytrainingselfsupervised} (LP, fine tuning the classifier head only\protect\footnotemark), FourierFT\cite{gaoParameterEfficientFineTuningDiscrete2024} ($n=\{3000,10000\}$), LoRA\cite{huLoRALowRankAdaptation2021} ($r=16$), and Full Fine-Tuning.  \loldu$_{r=768}$ outperforms LoRA$_{r=16}$ with 96.837\% fewer trainable parameters. Particularly noteworthy is that \loldu with $r=1$ achieves competitive scores with just 24 trainable parameters, while \loldu with $r=768$ attains the highest accuracy: 42.15\% for FGVC and 66.66\% for StanfordCars, showcasing the scalability and effectiveness of our approach. Full Fine-Tuning (85.8M parameters) and Linear Probing represent the upper and lower performance bounds, respectively.}
    \vspace{-2em}
    \label{fig:cv1}
\end{figure}

\footnotetext{Kindly note that the parameter count reported does not include the classification head, as it must be trained using all methods.}

Despite advancements in PEFT, the process of fine-tuning large models remains prohibitively expensive in terms of both computational resources and storage requirements. For instance, fine-tuning a model with 7 billion parameters, such as LLaMA2\cite{touvronLlamaOpenFoundation2023}, on instruct-following tasks\cite{alpaca, vicuna2023} incurs substantial costs. These costs are not limited to the training phase but extend to the storage of multiple fine-tuned model checkpoints, each consuming gigabytes of storage, thus leading to significant storage overhead. Approaches like Low-Rank Adaptation (LoRA)\cite{huLoRALowRankAdaptation2021} and Vector-based Random Matrix Adaptation (VeRA)\cite{kopiczko2024vera} have been developed to address these challenges by reducing the number of updated parameters. LoRA\cite{huLoRALowRankAdaptation2021} achieves this by randomly initializing two low-rank matrices and optimizing them to approximate the model's updated weights. Similarly, VeRA\cite{kopiczko2024vera} involves the random initialization and freezing of two matrices while training only two vectors for scale transformation. Recent research has revealed LoRA's limitations in data memorization due to low-rank updates. MoRA\cite{jiangMoRAHighRankUpdating2024} addresses this issue through input dimension reshaping and square linear layer application. However, these methods often result in suboptimal convergence due to random initialization, as proposed by \cite{pmlr-v9-glorot10a, pmlr-v28-sutskever13}, thus yielding a provably small hyperspherical energy \cite{qiuControllingTexttoImageDiffusion2024}. Furthermore, there is an accuracy gap compared to full fine-tuning, underscoring the need for more effective Parameter-Efficient Fine-Tuning strategies.

Thus, OFT \cite{qiuControllingTexttoImageDiffusion2024} proposes that maintaining orthogonality is crucial for preserving pre-trained knowledge, which enhances generalization \cite{liuOrthogonalOverParameterizedTraining2021}.Building on this insight, we observe that Lower-Diag-Upper (LDU) decomposition inherently possesses orthogonal properties in its lower and upper triangular matrices. Additionally, we incorporate a heuristic initialization constrain the range of initialized values, resulting in a more stable training process.

In contrast to other PEFT approaches \cite{huLoRALowRankAdaptation2021, mengPiSSAPrincipalSingular2024, kopiczko2024vera, jiangMoRAHighRankUpdating2024}, which require fine-tuning $O(n^2)$ level parameters, for the first time, we demonstrate that it is possible to optimize only 0.00025\% of parameters without any performance degradation. Our method, \loldunospace, operates at $O(n)$ level and employs the LDU decomposition technique to extract the core model parameters, which are then fine-tuned for downstream tasks.

To demonstrate the efficiency of \loldu across various model architectures, scales, and task types, we conduct an extensive set of experiments on tasks including  instruction following\cite{alpaca,vicuna2023, chia2023instructevalholisticevaluationinstructiontuned}, natural language understanding (NLU)\cite{wangGLUEMultiTaskBenchmark2019},  image classification \cite{cifar10,cifar100,flowers,cars,fgvc,eurosat}, and image generation\cite{ruiz2023dreamboothfinetuningtexttoimage}. These experiments involved models with architectures such as LLaMA2-7B (decoder-only)\cite{touvronLlamaOpenFoundation2023}, RoBERTa-Base (encoder-decoder)\cite{liuRoBERTaRobustlyOptimized2019}, ViT-Base (encoder-only)\cite{dosovitskiy2021imageworth16x16words}, and Stable Diffusion \cite{rombach2022highresolutionimagesynthesislatent}, with model scales ranging from 86 million to 7 billion parameters. This comprehensive evaluation verifies the effectiveness of our method across diverse scenarios.

In summary, this paper makes three key contributions:
\begin{itemize}
    \item We introduce a novel approach to Parameter-Efficient Fine-Tuning (PEFT) by firstly attempting to leverage Lower-Diag-Upper (LDU) decomposition, offering a solution that maintains model performance while drastically reducing trainable parameters to as low as 0.00025\% of the original model.
    
    \sloppy    
    \item We present \loldunospace, a PEFT technique that harnesses Low-Rank Adaptation via Lower-Diag-Upper Decomposition, which operates with a complexity of $O(n)$. The \loldu method employs orthogonal lower and upper triangular matrices to preserve pre-trained knowledge and enhance generalization, incorporating a heuristic initialization and scaling factor to optimize the diagonal matrix.

    \item  \loldu demonstrates the effectiveness and versatility through comprehensive experiments across various model architectures, scales, and task types. It offers a pioneering approach for efficient model adaptation across diverse scenarios in both NLP and CV domains.
\end{itemize}

\section{Related Work}
\label{sec:related_work}
\begin{figure}[t]
    \includegraphics[width=0.5\textwidth]{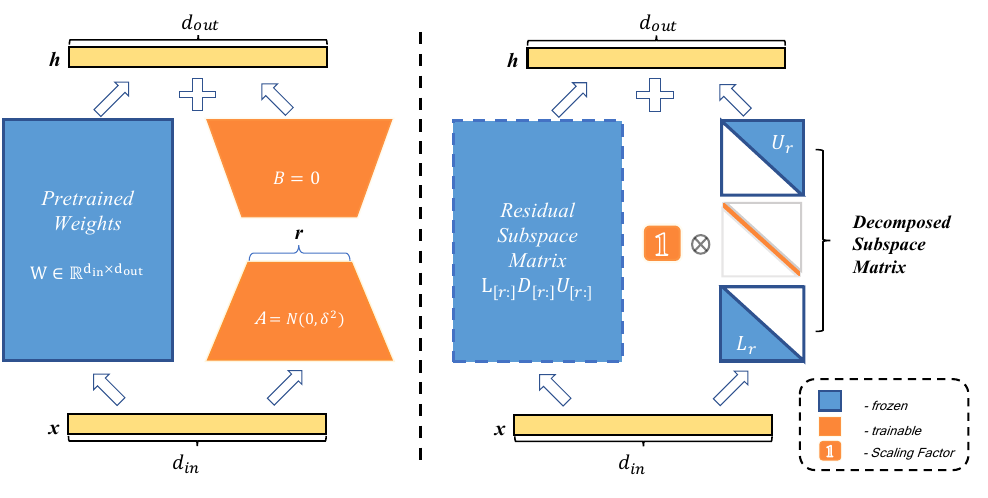}
    \caption{\textbf{Comparison of LoRA (left) and our \loldu (right) method.} In LoRA, tunable parameters are low-rank ($r$) matrices $A$ and $B$, with $\Delta W = BA$. For each weight $W$, there are $r \times (d_{in} + d_{out})$ trainable parameters. \loldunospace, however, optimizes a diagonal matrix for scale transformation, preserving original model knowledge during tuning. The weight update in \loldu is $\Delta W = \sigma \cdot P \cdot (L_r, \text{diag}(z_r), U_r)$, involving $r+1$ trainable parameters.  The permutation matrix $P$, while omitted in this figure for simplicity, is included in Figure \ref{fig:model}}
    \label{fig:compared_to_lora}
\end{figure}

Parameter-Efficient Fine-Tuning (PEFT) is designed to mitigate the significant computational and storage costs associated with Full Fine-Tuning (FT). Among the various PEFT approaches, Low-Rank Adaptation (LoRA) \cite{huLoRALowRankAdaptation2021} offers a more flexible and generalized re-parameterization framework for fine-tuning, achieved by training two low-rank matrices to approximate the updated parameters. However, studies\cite{pmlr-v9-glorot10a, pmlr-v28-sutskever13} have indicated that random initialization for re-parameterization can be a bottleneck, leading to suboptimal convergence. In this work, we present the first attempt to address this issue by leveraging the Lower-Diag-Upper (LDU) decomposition technique for initialization. In Figure \ref{fig:compared_to_lora}, we provide a comparison between LoRA and our \loldu method.

\textbf{Parameter efficient fine tuning (PEFT).} 
To date, existing PEFT approaches can be divided into three categories: (1) \textbf{Additive PEFT}: This approach introduces new tunable parameters or modifies model representations. Examples include adapters\cite{houlsbyParameterEfficientTransferLearning2019,leiConditionalAdaptersParameterefficient2023,zhangLLaMAAdapterEfficientFinetuning2023,zhangSpectralAdapterFineTuning2024} and prefix-tuning\cite{liPrefixTuningOptimizingContinuous2021}, which add small, trainable components to the model for efficient task-specific learning.  (2)  \textbf{Selective PEFT}\cite{guoParameterEfficientTransferLearning2021,dasUnifiedLowResourceSequence2023,ansellScalingSparseFineTuning2024,sungTrainingNeuralNetworks2021}: This method fine-tunes only a subset of the model's parameters, such as specific layers or neurons. Techniques like BitFit\cite{zakenBitFitSimpleParameterefficient2022} aims to only update bias parameters $b$, while maintaining fixed weights $W$, to shift the model's conditional distribution $p(y|x;\theta)$ towards the target domain distribution $p_\text{target}(y|x)$, where $\theta$ denotes the model parameters. (3) \textbf{Re-parameterized PEFT} \cite{huLoRALowRankAdaptation2021,liuDoRAWeightDecomposedLowRank2024,9662281}.  This technique usually reconstructs model parameters in a low-dimensional space as new knowledge is often represented in a low-rank form \cite{aghajanyanIntrinsicDimensionalityExplains2020}.

\textbf{Low-Rank Adaptation.}
LoRA\cite{huLoRALowRankAdaptation2021} decomposes parameter matrices into low-rank forms, maintaining performance while reducing the number of parameters to be fine-tuned. Previous studies have credited LoRA for its efficiency in inference and storage, albeit at an expensive training cost due to the random initialization, which causes the model to saturate more slowly. 
Recent studies \cite{phang2022hypertuningadaptinglargelanguage} have attempted to bridge this gap by exploring the development of new initialization methods to create LoRA parameters instead of starting from scratch. 
Advancing the initialization strategies for LoRA parameters is imperative for enhancing the quality and adaptability of downstream tasks. Therefore, Section \ref{sec:experiments} delves into the exploration of various initialization methodologies.

\textbf{Re-parameterization.}
Singular Value Decomposition (SVD) is widely utilized for re-parameterization in Parameter-Efficient Fine-Tuning (PEFT) methods. Recent studies \cite{mengPiSSAPrincipalSingular2024,fengTriLoRAIntegratingSVD2024,lingamSVFTParameterEfficientFineTuning2024,zhangSpectralAdapterFineTuning2024,han2023svdiffcompactparameterspace} have explored various SVD-based approaches for low-rank matrix initialization. These include fine-tuning singular values of reshaped weight matrices \cite{han2023svdiffcompactparameterspace}, initializing adapter matrices with principal components \cite{mengPiSSAPrincipalSingular2024}, introducing intermediate matrices between frozen principal components matrices, and updating weights as sparse combinations of singular vector outer products \cite{lingamSVFTParameterEfficientFineTuning2024}. However, SVD's computational complexity $O(mn^2+n^3)$ for an $m\times n$ matrix remains a constraint compared to LDU decomposition $O(mn^2-n^3/3)$. Furthermore, LDU decomposition offers a more interpretable representation of matrix structure through elementary row operations and pivoting strategies.

\section{Method}
\label{sec:method}

\begin{figure*}[t]
    \centering
    \includegraphics[width=1\textwidth]{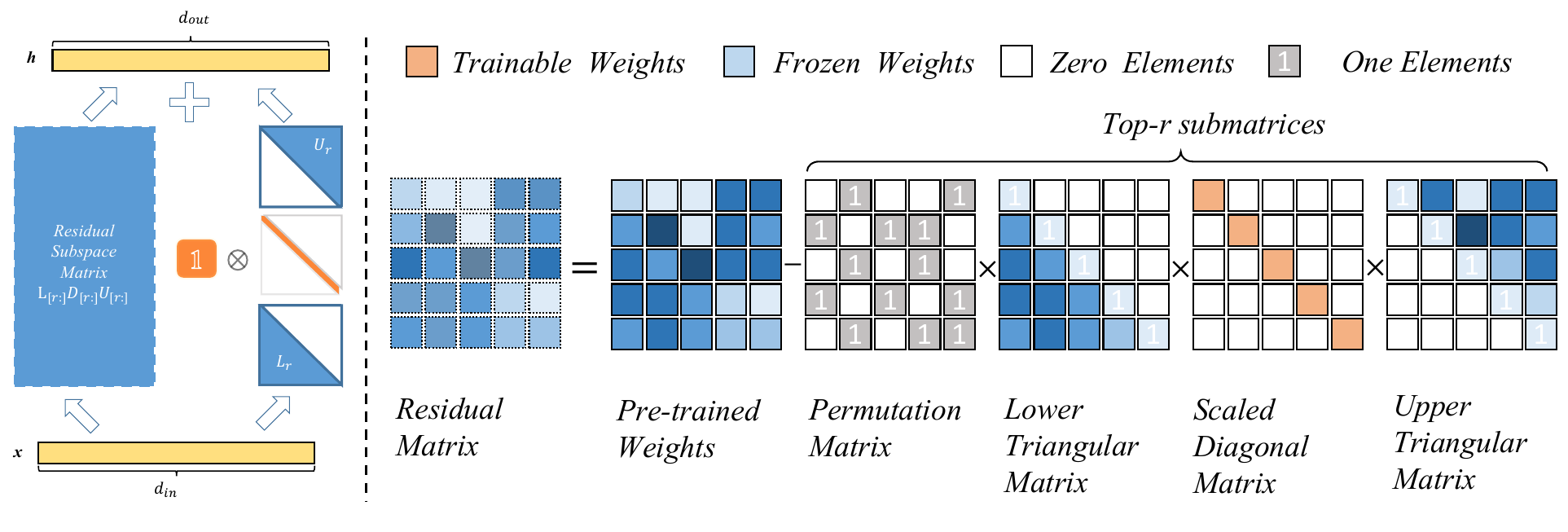}
    \caption{\textbf{Schematic representation of our \loldu method.} The left diagram illustrates the forward pass, demonstrating the transformation of the input  $x \in \mathbb{R}^{d_{in}}$ into the output $h \in \mathbb{R}^{d_{out}}$ via a residual subspace matrix $L_{[r:]}D_{[r:]}U_{[r:]}$ and a decomposed subspace matrix $\sigma L_rD_rU_r$. The right diagram shows the initialization process, where the residual matrix is obtained by performing LDU decomposition on the pre-trained weights, then subtracting the top-$r$ submatrices (top-$r$ rows and columns) from the permutation matrix (P), lower triangular (L), scaled diagonal (D), and upper triangular (U) matrices. Diagonal matrix is trainable (orange), while the other matrices remain fixed  (blue). \loldu enables efficient adaptation of pre-trained models via low-rank updates, reducing both computational cost and parameter count.}
    \label{fig:model}
\end{figure*}

We present \loldu (depicted in Figure~\ref{fig:model}), a parameter-efficient-fine-tuning method utilizing Lower-Diag-Upper (LDU) decomposition. \loldu builds upon the principle proposed by LoRA~\cite{huLoRALowRankAdaptation2021}, focusing on learning the changes in pre-trained weights. In contrast to LoRA, which employs random initialization, \loldu leverages the LDU decomposition for initialization. We then compute the Residual Subspace Matrix (RSM) by applying element-wise subtraction of the Decomposition Subspace Matrix (DSM) from the original matrix. The DSM is constructed using the first $r$ entries, which are selected to maintain a low-rank formation while remaining trainable.
\subsection{Initialization and Orthogonal Space Preservation}
Previous works have shown that maintain the orthogonality nature is crucial to improve the representation quanlity~\cite{qiuControllingTexttoImageDiffusion2024}. The advantage of LDU decomposition is the factorization that preserves the orthogonality of the lower and upper triangular matrices. We leverage this property to initialize the low-rank matrices. The LDU decomposition factorizes a matrix $W_0 \in \mathbb{R}^{m\times n}$ into four matrices:
\begin{equation}
    W_0 = P \cdot L \cdot \text{diag}(z) \cdot U,
    \label{eq:1}
\end{equation}
where $P \in \mathbb{R}^{m\times m}$ is a permutation matrix, $L \in \mathbb{R}^{m\times k}$ is lower triangular with ones on the diagonal, $\text{diag}(z) \in \mathbb{R}^{k\times k}$ is the diagonal formation of vector $z$, and $U \in \mathbb{R}^{k\times n}$ is upper triangular with ones on the diagonal, where $k=min(m,n)$. This property is essential for obtaining an equivalent formation to the original model weight $W_0$. 

Specifically, we optimize only the diagonal entries of matrix $\text{diag}(z)$ and dynamically adjust the scaling factor $\sigma$ to align updated parameters with the target matrix, wherein the $\sigma$ is initialized to 1.0.

\subsection{Low-Rank Approximation}
In the realm of learning weight changes, our approach aligns with the principles of LoRA-based methods~\cite{huLoRALowRankAdaptation2021, jiangMoRAHighRankUpdating2024, dettmersQLoRAEfficientFinetuning2023,zhangAdaLoRAAdaptiveBudget2023}, which mitigate inference latency by merging pre-trained weights with the learned adapter matrices. 

Formally, let $ W_0 \in \mathbb{R}^{m \times n} $ represent the pre-trained weight matrix, and $ \Delta W \in \mathbb{R}^{m \times n} $ denote the weight changes introduced during fine-tuning. LoRA parameterizes $ \Delta W $ using a low-rank decomposition in the forward pass:
\begin{equation}
    h = W_0 x + \Delta W x = W_0 x + B A x,
\end{equation}
where $ B \in \mathbb{R}^{m \times r} $ and $ A \in \mathbb{R}^{r \times n} $ are trainable matrices, with the rank $ r \ll \min(m, n) $.

In contrast, our proposed method, \loldunospace, decomposes the weight matrix $ W_0 $ using an LDU (Lower-Diag-Upper) decomposition which breaks down \( W_0 \) into four matrices: $P \cdot L \cdot \text{diag}(z) \cdot U$. We take the insprition from \cite{li2018measuringintrinsicdimensionobjective,aghajanyanIntrinsicDimensionalityExplains2020} that learned adapter matrices reside in a low intrinsic dimension. Therefore, we extract the top \( r \) components from the LDU decomposition, which helps in maintaining an intrinsic subspace to adapt to downstream tasks. These components are represented as follows:
\begin{equation}
    B = L_r = L_{[:,:r]} \in \mathbb{R}^{m \times r}, 
\end{equation}
\begin{equation}
    \text{diag}(z_r) = D_{[:r,:r]} \in \mathbb{R}^{r \times r},
\end{equation}
\begin{equation}
    A = U_r = U_{[:r,:]} \in \mathbb{R}^{r \times n},
\end{equation}
where \( L_r \) represents the first \( r \) columns of the lower triangular matrix \( L \), \( D_{[:r,:r]} \) denotes the top \( r \) by \( r \) block of the diagonal matrix \( D \), and \( U_r \) is the first \( r \) rows of the upper triangular matrix \( U \). These components capture the essential structure of the original weight matrix in a reduced form.
\subsection{\loldu Weight Adaptation Procedure}
Using these components, we define the Decomposed Subspace Matrix (DSM), which reconstructs a part of the original weight matrix using the top \( r \) components. The DSM is formulated as:
\begin{equation}
    DSM = \sigma \cdot P \cdot (L_r, \text{diag}(z_r), U_r),
\end{equation}
where \( \sigma \) is introduced to control the magnitude of the weight updates as a scaling factor.

Next, we obtain the Residual Subspace Matrix (RSM) by subtracting the DSM from the original weight matrix \( W_0 \), which ensures that the RSM captures the information not represented by the top \( r \) components, thereby preserving the full knowledge encoded in \( W_0 \):
\begin{equation}
    RSM = W_0 - DSM.
\end{equation}

The weight change \( \Delta W \) is parameterized as:
\begin{equation}
    \Delta W = DSM = \sigma \cdot P \cdot (L_r, \text{diag}(z_r), U_r),
\end{equation}
by parameterizing $\Delta W$ in this manner, efficient updates to the model weights are enabled without significantly increasing the parameter count.

The advantage of \loldu lies in its use of orthogonal, lower, and upper triangular matrices, which help preserve the inherent knowledge of the model. The orthogonal nature of these matrices ensures that the decomposed components maintain their properties during transformations as proposed by \cite{liuOrthogonalOverParameterizedTraining2021}, thereby preserving the information integrity. 
Moreover, we initialize $diag(z_r)$ using heuristic methods such as Constant ($D_r.mean$), Uniform, Normal, or Regular LDU, to enhance training stability.

The proposed forward pass can be expressed as follows:

\begin{equation}
    \begin{split}
        h &= RSM x + \Delta W x  \\
          &= RSM x + DSM x \\
          &= RSM x + \sigma \cdot P \cdot (L_r, \text{diag}(z_r), U_r)x.
    \end{split}
\end{equation}
    
\begin{algorithm}
    \definecolor{mygray}{rgb}{0.5, 0.5, 0.5}
    \renewcommand{\algorithmiccomment}[1]{\hfill\textcolor{mygray}{// #1}}
    \renewcommand{\algorithmicrequire}{\textbf{Input:}}
    \renewcommand{\algorithmicensure}{\textbf{Output:}}
    \caption{Low-Rank LDU Decomposition and Optimization for Layer Weight Adaptation}
    \label{algo:loldu}
    \begin{algorithmic}[1]
    \Require Weight matrix $\mathbf{W} \in \mathbb{R}^{m \times n}$, rank $r$, alpha $\alpha$, learning rate $\eta$, number of iterations $T$, projection operator $\mathcal{P}$
    \Ensure Decomposed components $\mathbf{P}$, $\mathbf{L}_r$, $\mathbf{D}_r$, $\mathbf{U}_r$, residual $\mathbf{W}_\text{residual}$, scaling factor $\sigma$, optimized $D_r$, optimized $\sigma$
    
    \State \textbf{Phase 1: Initial Decomposition}
    \State $\mathbf{P}$, $\mathbf{L}$, $\mathbf{U} \gets \text{LU\_decomposition}(\mathbf{W})$ \Comment{Perform standard LU decomposition}
    \State $\mathbf{D} \gets \text{diag}(\mathbf{U})$ \Comment{Extract diagonal matrix}
    \State $\mathbf{U} \gets \mathbf{D}^{-1}\mathbf{U}$ \Comment{Normalize U}
    
    \State \textbf{Phase 2: Low-Rank Approximation}
    \State $\mathbf{L}_r \gets \mathbf{L}_{:,1:r}$ \Comment{Extract first $r$ columns of L}
    \State $\mathbf{D}_r \gets \mathbf{D}_{1:r,1:r}$ \Comment{Extract top-left $r \times r$ submatrix of D}
    \State $\mathbf{U}_r \gets \mathbf{U}_{1:r,:}$ \Comment{Extract first $r$ rows of U}
    
    \State \textbf{Phase 3: Scaling Factor and Residual Computation}
    \State $\sigma \gets \alpha / r$ \Comment{Compute scaling factor}
    \State $\mathbf{W}_\text{approx} \gets \sigma \mathbf{P}\mathbf{L}_r\mathbf{D}_r\mathbf{U}_r$ \Comment{Compute low-rank approximation}
    \State $\mathbf{W}_\text{residual} \gets \mathbf{W} - \mathbf{W}_\text{approx}$ \Comment{Compute residual matrix}
    
    \State \textbf{Phase 4: Heuristic Initialization}
    \State Apply heuristic initialization to $\mathbf{D}_r$ \Comment{Choose from methods: Constant($(D_r.\text{mean})$), Uniform, Normal, or Regular LDU}
    
    \State \textbf{Phase 5: Optimization with Projected Gradient Descent}
    \For{$t \gets 1$ to $T$}
        \State Compute gradients $\nabla_{D_r} \mathcal{L}$ and $\nabla_{\sigma} \mathcal{L}$
        \State $D_r \gets \mathcal{P}(D_r - \eta \cdot \nabla_{D_r} \mathcal{L})$
        \State $\sigma \gets \mathcal{P}(\sigma - \eta \cdot \nabla_{\sigma} \mathcal{L})$
    \EndFor

    \State \Return $\mathbf{P}$, $\mathbf{L}_r$, $\mathbf{D}_r$, $\mathbf{U}_r$, $\mathbf{W}_\text{residual}$, $\sigma$
    \end{algorithmic}
\end{algorithm}

\subsection{Optimization Process}

The fine-tuning phase of \loldu employs a sophisticated optimization strategy, focusing on the diagonal matrix $D_r$ and the scaling factor $\sigma$. This approach represents a departure from conventional fine-tuning methods, offering more granular control over parameter updates while preserving the integrity of pre-trained knowledge.

The optimization problem is formulated as a constrained minimization:

\begin{equation}
\begin{aligned}
    & \underset{D_r, \sigma}{\text{minimize}} \quad \mathcal{L}(f_{W_0 + \Delta W}(x), y) \\
    & \text{subject to} \quad \|D_r\|_F \leq \epsilon, \\
    & \phantom{\text{subject to}} \quad 0 < \sigma \leq 1,
\end{aligned}
\end{equation}
where $\mathcal{L}$ denotes the task-specific loss function, $f_{W_0 + \Delta W}$ denotes the model with updated weights, $(x, y)$ are the input-output pairs from the fine-tuning dataset, $\|\cdot\|_F$ represents the Frobenius norm, and $\epsilon$ is a set constraint threshold.

To address the constrained nature of the optimization problem, we employ a projected gradient descent method, ensuring that updates to $D_r$ and $\sigma$ remain within the feasible region defined by the constraints. This is achieved through a projection operator $\mathcal{P}$:

\begin{equation}
    D_r^{(t+1)} = \mathcal{P}\left(D_r^{(t)} - \eta_t \frac{\partial \mathcal{L}}{\partial D_r^{(t)}}\right),
\end{equation}

\begin{equation}
    \sigma^{(t+1)} = \mathcal{P}\left(\sigma^{(t)} - \eta_t \frac{\partial \mathcal{L}}{\partial \sigma^{(t)}}\right),
\end{equation}
where $\eta_t$ is the learning rate at iteration $t$, adaptively adjusted using techniques such as Adam\cite{kingma2017adammethodstochasticoptimization} or RMSprop\cite{RMSProp} to account for the geometry of the parameter space.

Please refer to Algorithm~\ref{algo:loldu} for additional detailed information. 

\subsection{Computational Complexity Analysis}
The computational efficiency of \loldu can be evaluated in terms of both space and time complexity:

\textbf{Space complexity:} The storage requirement for \loldu is $O(r+1)$, which is considerably lower than the $O(mr + rn)$ required by methods such as LoRA. This reduction in parameter count not only leads to significant memory savings but improves efficiency during both the training and inference phases.

\textbf{Time complexity:} The forward pass of \loldu requires $O(mnr)$ operations with a minor linear term $O(r)$. In contrast to methodologies that necessitate recurrent complex iterations \cite{ding2023sparselowrankadaptationpretrained, zhangAdaLoRAAdaptiveBudget2023}, \loldu performs the LDU decomposition only once during initialization, with a time complexity of $O(mn^2-n^3/3)$, and utilizing direct updates via projected gradient descent without iterative refinement, ensuring efficient parameter optimization and rapid convergence.

In summary, \loldu leverages LDU decomposition to efficiently parameterize weight changes, reducing the number of tunable parameters and maintaining high performance. This method provides a more efficiency and effective alternative to traditional LoRA-based approaches.

\section{Experiments}
\label{sec:experiments}
\begin{table*}[t]
    \centering
    \small
    \setlength{\tabcolsep}{6pt}
    \renewcommand{\arraystretch}{0.80}
    \caption{Results for different adaptation methods on the GLUE benchmark. The term "Params" refers to the number of parameters updated during fine-tuning. We report Matthew's correlation for CoLA, Pearson correlation for STS-B, and accuracy for the remaining tasks. Higher values indicate better performance. Except LoLDU, all results are from prior work. LoLDU performs on par with LoRA while using significantly fewer parameters. The $\Delta_{baseline}$ row shows the percentage increase or decrease in performance compared to our method.}
    \begin{tabular}{l|l|r|ccccccc}
        \toprule
        \multirow{2}{*}{Model} & \multirow{2}{*}{Method} & \parbox{1.5 cm}{\#  Params} & \multicolumn{1}{c}{SST-2} & \multicolumn{1}{c}{MRPC} & \multicolumn{1}{c}{CoLA} & \multicolumn{1}{c}{QNLI} & \multicolumn{1}{c}{RTE} & \multicolumn{1}{c}{STS-B} & \multicolumn{1}{c}{Avg.} \\
        & & &acc &acc &cor &acc &acc &cor & \\ \midrule
        \multirow{8}{*}{\rotatebox{90}{RoBERTa-Base}} 
        
        & FT & 125M & 94.8 & 90.2 & 63.6 & 92.8 & 78.7 & 91.2 & 85.2 \\
        & BitFit & 0.1M & 93.7 & \textbf{92.7} & 62.0 & 91.8 & 81.5 & 90.8 & 85.4 \\
        & LoRA & 0.3M & \textbf{95.1} & 89.7 & 63.4 & \textbf{93.3} & 78.4 & 91.5 & 85.2 \\
        & PiSSA & 0.707M & 94.6 & 88.4 & 63.0 & 93.1 & \textbf{85.9} & 91.2 & \textbf{86.0} \\
        & VeRA & 0.043M & 94.6 & 89.5 & \textbf{65.6} & 91.8 & 78.7 & 90.7 & 85.2 \\
        \cmidrule(l){2-10}
        & \cellcolor{gray!20}LoLDU & \cellcolor{gray!20}\textbf{0.0184M} & \cellcolor{gray!20}94.8 & \cellcolor{gray!20}89.9 & \cellcolor{gray!20}63.8 & \cellcolor{gray!20}92.9 & \cellcolor{gray!20}81.3 & \cellcolor{gray!20}\textbf{92.3} & \cellcolor{gray!20}85.8 \\
        & $\Delta_{baseline}$ & \textcolor{red}{\textbf{6.13\%}} & \textcolor{blue}{-0.3} & \textcolor{red}{\textbf{+0.2}} & \textcolor{red}{\textbf{+0.4}} & \textcolor{blue}{-0.4} & \textcolor{red}{\textbf{+2.9}} & \textcolor{red}{\textbf{+0.8}} & \textcolor{red}{\textbf{+0.6}} \\
        
        \midrule
    \end{tabular}
    \label{tab:results_glue}
\end{table*}

This section presents an evaluation of \loldu within the fields of natural language processing (NLP) and computer vision (CV).
For NLP, \loldu is applied for fine-tuning: (1) RoBERTa Base\cite{liuRoBERTaRobustlyOptimized2019} on natural language understanding (GLUE\cite{wangGLUEMultiTaskBenchmark2019}), and (2) LLaMA-2 7B\cite{touvronLlamaOpenFoundation2023} on instruction tuning (Alpaca \cite{alpaca}, Vicuna\cite{vicuna2023}). For CV, we apply \loldu to fine-tune: (1) Vision Transformers (ViT) Base \cite{dosovitskiy2021imageworth16x16words} on image classification \cite{cifar10,cifar100,flowers,cars,fgvc,eurosat}, and (2) Stable Diffusion v1.5 \cite{rombach2022highresolutionimagesynthesislatent} on customized image generation\cite{ruiz2023dreamboothfinetuningtexttoimage}.

We compare our \loldu method with widely used Parameter-Efficient Fine-Tuning (PEFT) methods. To ensure a fair comparison, we replicate the setups from previous studies \cite{huLoRALowRankAdaptation2021,gaoParameterEfficientFineTuningDiscrete2024,ren-etal-2024-melora} and utilize their reported results. 

The baselines considered are:
\begin{itemize}
    \item \textbf{Full Fine-Tuning (FT)}: FT trains all model parameters on the task-specific data.
    \item \textbf{LoRA}~\cite{huLoRALowRankAdaptation2021}: LoRA updates weights by injecting two tunable low-rank matrices for parameterization.
    \item \textbf{MELoRA}~\cite{ren-etal-2024-melora}: MELoRA trains a group of mini LoRAs to maintain a higher rank.
    \item \textbf{FourierFT}~\cite{gaoParameterEfficientFineTuningDiscrete2024}: FourierFT learns a small fraction of spectral coefficients using the Fourier transform.
\end{itemize}

Finally, we perform ablation studies to examine the impact of initialization methods, scaling factors, and rank. Further results concerning the learning rate and rank are detailed in Appendix \ref{app:lr} and Appendix \ref{app:rank}. We conduct all experiments on a single NVIDIA RTX A6000 (48G) GPU.

\begin{table}[!h]
    \centering
    \scriptsize
    \setlength{\tabcolsep}{1pt}
    \renewcommand{\arraystretch}{0.8} %
    \caption{ Comparative analysis of various methods on image classification datasets using ViT Base models. The table reports the mean accuracy (\%) after 10 epochs, alongside parameters efficiency and approach features.
    }

    \begin{adjustbox}{max width=\textwidth}
    \begin{tabular}{l|cccccc}
    \toprule
    
    \multirow{1}{*}{Method} & \multicolumn{1}{c}{\makecell{Mean \\Acc.}} & \multicolumn{1}{c}{\makecell{Params \\(\%)}} & \multicolumn{1}{r}{\makecell{Keep \\Orthogonal}} & \multicolumn{1}{l}{\makecell{No random \\ Init.}} & \multicolumn{1}{l}{\makecell{No extra \\ Infer. cost}} & \multicolumn{1}{l}{\makecell{Faster \\convergence}}  \\
    
    \midrule
    
    FullFT         & 88.20     &100   & \ding{55} &  \ding{51} & \ding{51} & \ding{51}  \\
    LP   & 68.38    &-   & \ding{55} & \ding{55} & \ding{51} & \ding{55} \\
    \midrule
    LoRA       & 76.22    &6.77   & \ding{55} & \ding{55} & \ding{51} & \ding{55} \\
    FourierFT     & 79.29    &2.79   & \ding{55} & \ding{55} & \ding{51} & \ding{55} \\
    \midrule
    \cellcolor{gray!20}\loldu      & \cellcolor{gray!20}82.79    &\cellcolor{gray!20}0.21   & \cellcolor{gray!20}\ding{51} & \cellcolor{gray!20}\ding{51} &\cellcolor{gray!20}\ding{51}& \cellcolor{gray!20}\ding{51}\\
    
    \bottomrule
    \end{tabular}
    \end{adjustbox}

    \label{tab:comparision}
\end{table}

\subsection{Natural Language Understanding}
\paragraph{Models and Datasets}
We evaluate \loldu on the GLUE benchmark (General Language Understanding Evaluation~\cite{wangGLUEMultiTaskBenchmark2019}), which comprises nine NLU tasks. These tasks include single-sentence classification (CoLA, SST-2), similarity and paraphrasing (MRPC, STS-B, QQP), and natural language inference (MNLI, QNLI, RTE, WNLI). For evaluation, we fine-tune pre-trained RoBERTa Base models~\cite{liuRoBERTaRobustlyOptimized2019}.

\paragraph{Implementation Details}
We adopt the experimental setup of VeRA \cite{kopiczko2024vera}, tuning the hyperparameters for learning rates and the scaling factor values across six datasets in the GLUE benchmark. Following the approach of LoRA \cite{huLoRALowRankAdaptation2021}, we fully fine-tune the classification head. We apply \loldu to the weight matrices  $W_q$, $W_k$, $W_v$, and $W_o$ in each transformer block. Hyperparameters are provided in Table~\ref{tab:hp_glue} in the Appendix.

\paragraph{Results}
Results are summarized in Table~\ref{tab:results_glue}. Following \cite{huLoRALowRankAdaptation2021}, \cite{zhangAdaLoRAAdaptiveBudget2023}, and \cite{valipourDyLoRAParameterEfficient2023}, we specify the number of trainable parameters for the fine-tuned layers excluding the classification head. We report the median of five random seed results, selecting the best epoch for each run. In general, \loldu achieves better or on-par performance compared to baseline methods with significantly fewer trainable parameters. Notably, \loldu outperforms all baselines including fully fine-tuning the RoBERTa Base on STS-B. As mentioned in Section~\ref{sec:method}, the parameter count of LoRA is dependent on both the width and depth of models, resulting in a larger count growth (LoRA: $0.3\text{M}$; ours: $0.0184\text{M}$) compared to \loldunospace.

\subsection{Instruction Tuning}
\paragraph{Models and Datasets}
Instruction tuning \cite{longpre2023flancollectiondesigningdata,köpf2023openassistantconversationsdemocratizing,alpaca} is a technique that involves fine-tuning large language models (LLMs) on paired data consisting of instructions and their corresponding outputs to enhance the quality of the model's responses. In our study, we apply LoRA \cite{huLoRALowRankAdaptation2021} and \loldu to fine-tune the LLaMA2 model \cite{touvronLlamaOpenFoundation2023}. Specifically, we use LLaMA2-7B as the base model, which is then fine-tuned on the Alpaca dataset \cite{alpaca}. This dataset comprises 52,000 instruction-output pairs generated by OpenAI's text-davinci-003 model. For evaluation, we conduct a rigorous and holistic assessment of the fine-tuned model using INSTRUCTEVAL \cite{chia2023instructevalholisticevaluationinstructiontuned}, allowing us to systematically analyze the model's performance in problem-solving, writing ability, and alignment to human values.

\paragraph{Implementation Details} In the implementation of LoRA, a rank of $r=64$ is employed, with a focus on updating all linear layers, excluding the language modeling head (\texttt{lm\_head}), and specifically targeting the $W_Q$ and $W_V$ matrices. For \loldunospace, the training process spans three epochs, and we present the average performance scores across all evaluated responses. Hyperparameter configuration is detailed in Table~\ref{tab:hp_llama} in Appendix \ref{app:hp_llama}.

\paragraph{Results} The results, as presented in Table \ref{tab:result_llama}, demonstrate that \loldu achieves a slight improvement over the performance of LoRA, while employing merely 0.05\% of the parameters required by LoRA.

\begin{table}[!h]
  \centering
  \footnotesize
  \setlength{\tabcolsep}{6pt}
  \caption{ 
    Results on INSTRUCTEVAL for instruction-following tasks: exact match for MMLU, DROP, and BBH, pass@1 for HumanEval. Higher values are preferable. Boldface indicates the best metric values. The $\Delta_{baseline}$ row displays the performance change percentage compared to our method.
  }
  \begin{tabular}{@{}l|l@{}|rcccc@{}}
      \toprule
       Model & Method & \# Params & MMLU & DROP & HEval & BBH\\ 
      \midrule 
      \multirow{7}{*}{\rotatebox{90}{LLaMA2-7B}} 
      & w/o FT & -  & 45.96 & 31.55 & 12.20 & 32.04 \\ 
      & LoRA  & 33.6M & 45.64 & 32.46 & 15.09 & 32.40 \\ 
      & AdaLoRA & 33.6M & 45.96 & 31.94 & 14.02 & 32.85 \\ 
      & MELoRA ~& 0.5M & \textbf{46.46} & 32.65 & \textbf{16.16} & 33.01 \\  
      \cmidrule(l){2-7}
      & \cellcolor{gray!20}LoLDU ~& \cellcolor{gray!20}\textbf{0.016M} & \cellcolor{gray!20}46.21 & \cellcolor{gray!20}\textbf{32.71} & \cellcolor{gray!20}15.11 & \cellcolor{gray!20}\textbf{33.12} \\  
      & $\Delta_{baseline}$ & \textcolor{red}{\textbf{0.05\%}} & \textcolor{red}{\textbf{+0.57}} & \textcolor{red}{\textbf{+0.25}} & \textcolor{red}{\textbf{+0.02}} & \textcolor{red}{\textbf{+0.72}} \\
      \bottomrule
  \end{tabular}
  \label{tab:result_llama}
\end{table}

\begin{figure*}[!h]
    \centering
    \vspace{-1em}
    \begin{adjustbox}{max width=\textwidth} %
        \includegraphics[width=1\textwidth]{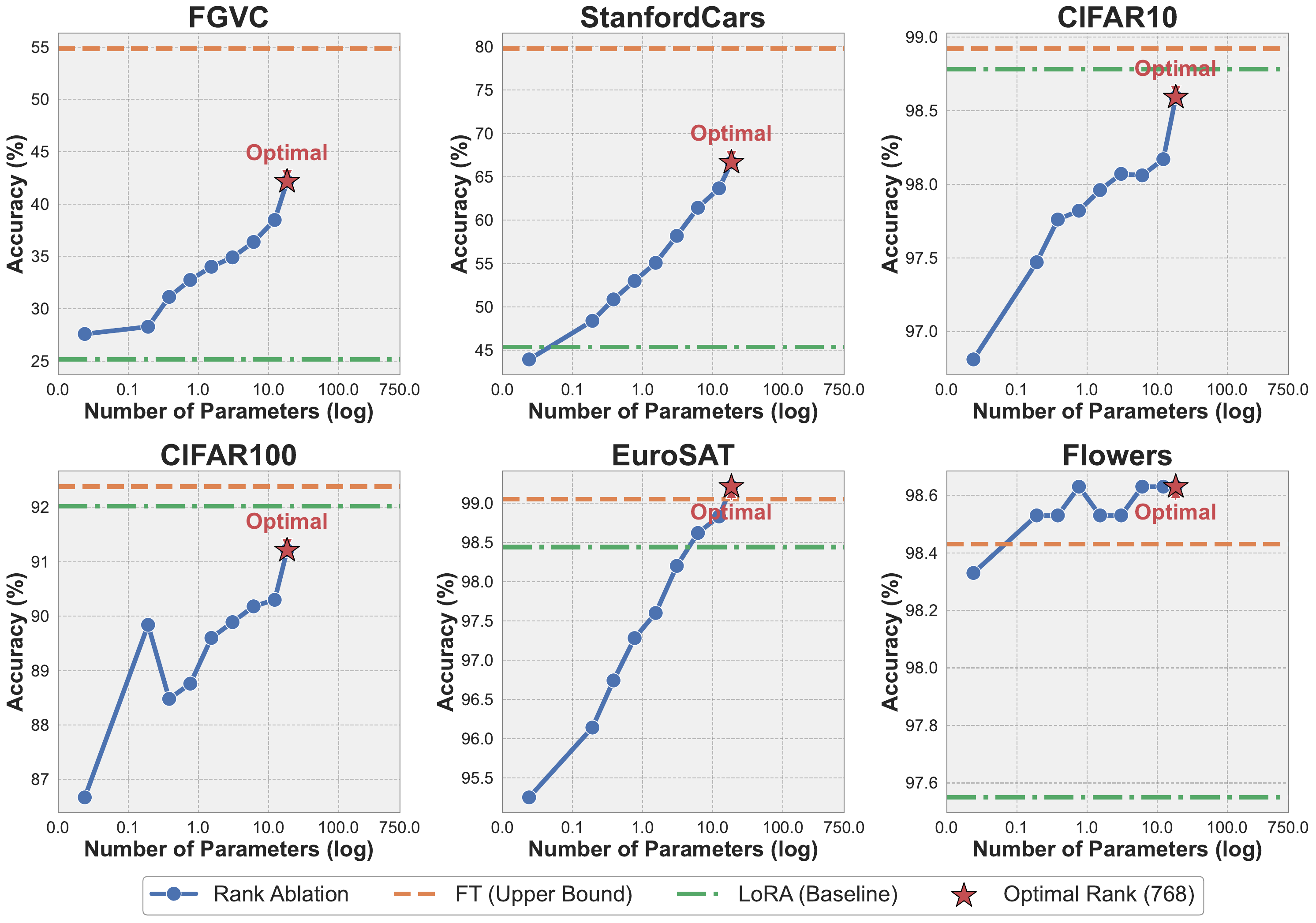}
    \end{adjustbox}
    \caption{\textbf{Comprehensive Analysis of Rank Ablation Study Results.} This figure presents the performance of the ViT-base model on various image classification tasks using the LoLDU method with different ranks. The x-axis shows ranks (1 to 768), and the y-axis indicates accuracy for datasets: FGVC, StanfordCars, CIFAR10, CIFAR100, EuroSAT, and Flowers.}
    \label{fig:rank}
\end{figure*}

\subsection{Image Classification}
\paragraph{Models and Datasets} 
We assess our approach on image classification utilizing the Base version of the Vision Transformer (ViT) \cite{dosovitskiy2021imageworth16x16words}, pre-trained on ImageNet-21K \cite{ridnikImageNet21KPretrainingMasses2021}. Fine-tuning is performed on datasets such as CIFAR10 (10) \cite{cifar10}, EuroSAT (10) \cite{eurosat}, as well as StanfordCars (196) \cite{cars}, FLOWERS102 (102) \cite{flowers}, FGVC (100) \cite{fgvc}, and CIFAR100 (100) \cite{cifar100}, covering both small and large label spaces. For detailed information, refer to Appendix \ref{app:cv1}.
\paragraph{Implementation Details} 
We include three baselines for evaluation: Full Fine-Tuning (FT), Linear Probing\cite{chen2021empiricalstudytrainingselfsupervised} (LP, fine-tuning the classification head only), and LoRA\cite{huLoRALowRankAdaptation2021}. We adhere to the experimental configurations established by FourierFT \cite{gaoParameterEfficientFineTuningDiscrete2024}. For both LoRA and our method, only the $W_Q$ and $W_V$ matrices of ViT are updated. We use $r=16$ for LoRA and $r=\{64, 768\}$ for \loldunospace. Detailed hyperparameter configurations are available in Table \ref{tab:hp_vit} in the Appendix \ref{app:cv1}.

\begin{table*}[t]
    \centering
    \small
    \setlength{\tabcolsep}{6pt}
    \setstretch{0.80}
    \caption{We conducted a comparison on image classification datasets using ViT Base models. The accuracy (\%) after 10 epochs is reported. FourierFT was evaluated using different trainable parameters for each layer, indicated by symbols: (\ding{62}) for 3000 and (\ding{61}) for 10000. $\Delta_{baseline}$ represents the performance gap between our LoLDU method and the baseline method LoRA. \textbf{Bold} denotes the best results.}
    \begin{adjustbox}{max width=\textwidth}
    \begin{tabular}{l|l|r|cccccc|c}
    \toprule
    Model & Method & \parbox{1.5 cm}{\# Params} & FGVC & StanfordCars & CIFAR10 & CIFAR100 & EuroSAT & Flowers & Avg. \\ 
    & & &acc &acc &acc &acc &acc &acc & \\ \midrule
    \multirow{10}{*}{\rotatebox{90}{ViT-Base}} 
    & LP & - & 17.44 & 25.76 & 96.41 & 84.28 & 88.72 & 97.64 & 68.38 \\
    & FT & 85.8M & \textbf{54.84} & \textbf{79.78} & \textbf{98.92} & \textbf{92.38} & 99.05 & 98.43 & \textbf{87.23} \\
    & LoRA(r16) & 581K & 25.16 & 45.38 & 98.78 & 92.02 & 98.44 & 97.55 & 76.22 \\
    & FourierFT(\ding{62}) & 72K & 27.51 & 46.11 & 98.58 & 91.20 & 98.29 & 98.14 & 76.64 \\
    & FourierFT(\ding{61}) & 239K & 32.44 & 56.36 & 98.69 & 91.45 & 98.78 & 98.04 & 79.29 \\    
    \cmidrule(l){2-10}
    & LoLDU(r64) & 1.5k & 32.31 & 50.99 & 97.96 & 89.60 & 97.60 & 98.53 & 77.83 \\
    & \cellcolor{gray!20}LoLDU(r768) & \cellcolor{gray!20}18k & \cellcolor{gray!20}42.15 & \cellcolor{gray!20}66.66 & \cellcolor{gray!20}98.59 & \cellcolor{gray!20}91.21 & \cellcolor{gray!20}\textbf{99.21} & \cellcolor{gray!20}\textbf{98.92} & \cellcolor{gray!20}82.79 \\
    								
    &$\Delta_{baseline}$  & \textcolor{red}{\textbf{3.173\%}} & \textcolor{red}{\textbf{+16.99}} & \textcolor{red}{\textbf{+21.28}} & \textcolor{blue}{-0.19} & \textcolor{blue}{-0.81} & \textcolor{red}{\textbf{+0.77}} & \textcolor{red}{\textbf{+1.37}} & \textcolor{red}{\textbf{+6.57}} \\
    
    \bottomrule
    \end{tabular}
    \end{adjustbox}
    \label{tab:cv}
\end{table*}

\paragraph{Results} Table \ref{tab:cv} presents the results for six image classification datasets using the ViT Base model. LoRA and \loldu demonstrate superior performance compared to Linear Probing \cite{chen2021empiricalstudytrainingselfsupervised}, showcasing their efficacy in image classification tasks within the computer vision domain. Notably, our approach achieves comparable outcomes while utilizing merely 3.173\% of LoRA's parameters. \loldu exhibits particularly impressive gains, surpassing LoRA by 15.28\% and 16.99\% in FGVC and StanfordCars tasks, respectively, effectively narrowing the accuracy gap with Full Fine-Tuning, as depicted in Figure \ref{fig:cv1}. Furthermore, \loldu outperforms all baselines, including Fully Fine-Tuning, on EuroSAT and Flowers datasets.

\subsection{Image Generation}

\begin{figure}[!h]
    \centering
    \vspace{-1em}
    \begin{adjustbox}{max width=\textwidth} %
        \includegraphics[width=0.5\textwidth]{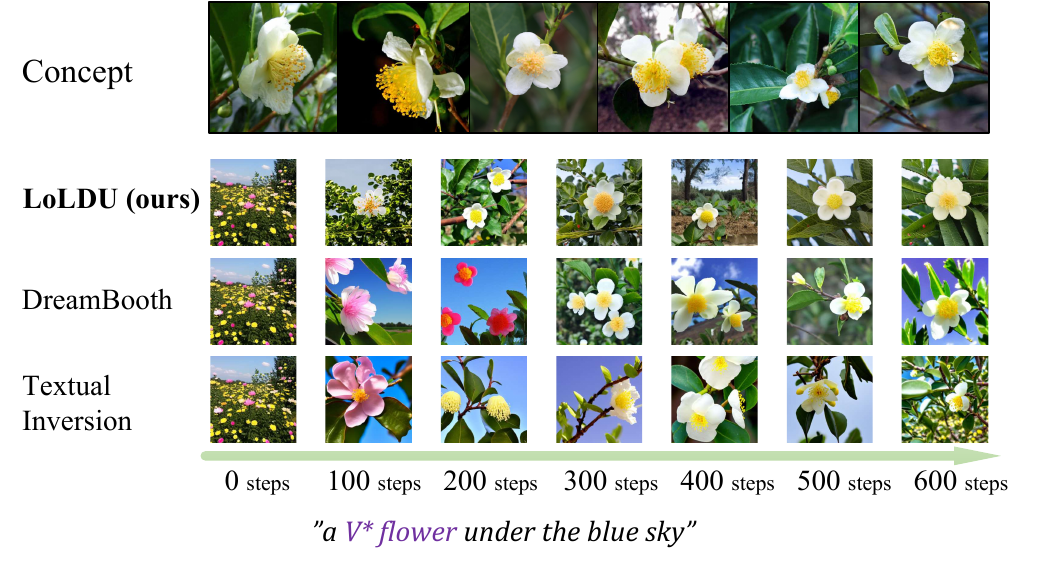}
    \end{adjustbox}
    \caption{\textbf{Concept Learning Progression In Text-to-Image Generation.}
    Top row: target concept. Subsequent rows: generated images using \loldu (our method), DreamBooth\cite{ruiz2023dreamboothfinetuningtexttoimage}, and Textual Inversion\cite{gal2022imageworthwordpersonalizing}, respectively, at training steps 0-600. \loldu exhibits accelerated convergence, achieving concept acquisition within $\sim$ 100 steps, surpassing baseline methods in efficiency.
    }
    \label{fig:cv2-1}
\end{figure}

\paragraph{Models and Datasets}We assess our method in the domain of image generation. Recent research \cite{gal2022imageworthwordpersonalizing,ruiz2023dreamboothfinetuningtexttoimage} highlights the necessity for customization in this field, which holds significant practical implications. The goal is to fine-tune a text-to-image model using a limited set (typically 3-5) of images representing an unique concept (e.g., a scene, individual, pet, or object) to effectively capture and reproduce the novel concept.
For this study, we employ the v1.5 version of Stable Diffusion (SD) \cite{rombach2022highresolutionimagesynthesislatent}, a widely-adopted computer vision foundation model. SD is pre-trained on LAION-5B \cite{schuhmann2022laion5bopenlargescaledataset}, a dataset consists of 5.85 billion image-text pairs filtered using CLIP \cite{radford2021learningtransferablevisualmodels}.

\paragraph{Implementation Details} 
We conduct our experiments on seven different concepts, including persons, pets, and objects, using the CustomConcept101 dataset \cite{kumari2023multiconceptcustomizationtexttoimagediffusion} and the human-centric FFHQ dataset \cite{ffhq}.
We select two concurrent works as baselines: Textual Inversion \cite{gal2022imageworthwordpersonalizing} and DreamBooth \cite{ruiz2023dreamboothfinetuningtexttoimage}. Textual Inversion learns new concept by mapping it from the image to the textual modality, encoding them as a rare token in the embedding space. DreamBooth, utilizes a semantic prior (e.g., class-specific) to maintain the subject's key features.
We provide the datasets in Figure \ref{fig:cv2} and hyperparameters in Table \ref{tab:hp_sd} in Appendix \ref{app:cv2}.

\begin{table*}[t]
    \centering
    \small
    \setlength{\tabcolsep}{4pt}
    \setstretch{1.2}
    \caption{Ablation study of different initialization methods across six image classification datasets. We set rank up to 768 and learning rate to 3e-3 and test on the ViT base model. The datasets include FGVC, StanfordCars, CIFAR10, CIFAR100, EuroSAT, and Flowers. The uniform initialization method is indicated by symbols: \ding{62} for (a=-1, b=1) and \ding{61} for (a=-z.mean/2, b=z.mean/2). The normal initialization method is indicated by symbols: \ding{58} for (mean=0, std=1) and \ding{72} for (mean=z.mean, std=z.std). For each entry, the left value represents results with scaling factor, while the right value in \textcolor{gray}{gray} represents results without scaling factor. The average performance (Avg.) across all datasets is also reported. \textbf{Bold} denotes the best results for each dataset and the average.}
    \begin{adjustbox}{max width=\textwidth}
    \begin{tabular}{ll|ccccccc}
    \toprule
    & \multirow{1}{*}{\makecell{Initialization\\ Method}} & \multicolumn{1}{c}{FGVC} & \multicolumn{1}{c}{StanfordCars} & \multicolumn{1}{c}{CIFAR10} & \multicolumn{1}{c}{CIFAR100} & \multicolumn{1}{c}{EuroSAT} & \multicolumn{1}{c}{Flowers} & \multicolumn{1}{c}{Avg.} \\
    & &acc &acc &acc &acc &acc &acc & \\ \midrule
    \multicolumn{8}{c}{\textbf{ViT-Base Initialization Ablation Study}} \\
    \midrule
    & Uniform(\ding{61}) & 2.37 / \textcolor{gray}{2.37} & 1.17 / \textcolor{gray}{1.38} & 35.92 / \textcolor{gray}{28.93} & 14.22 / \textcolor{gray}{9.71} & 57.81 / \textcolor{gray}{52.95} & 4.51 / \textcolor{gray}{4.41} & 19.33 / \textcolor{gray}{16.63} \\
    & Normal(\ding{58}) & 39.60 / \textcolor{gray}{39.12} & 65.17 / \textcolor{gray}{\textbf{65.00}} & 98.02 / \textcolor{gray}{\textbf{98.33}} & 90.27 / \textcolor{gray}{90.54} & 99.00 / \textcolor{gray}{\textbf{99.03}} & 98.63 / \textcolor{gray}{98.63} & 81.78 / \textcolor{gray}{81.78} \\
    & Normal(\ding{72}) & 2.10 / \textcolor{gray}{2.13} & 1.34 / \textcolor{gray}{1.12} & 29.17 / \textcolor{gray}{26.54} & 10.11 / \textcolor{gray}{7.91} & 52.98 / \textcolor{gray}{48.49} & 4.61 / \textcolor{gray}{4.41} & 16.72 / \textcolor{gray}{15.10} \\
    & Constant(z.mean) & \textbf{42.21} / \textcolor{gray}{\textbf{41.16}} & 65.41 / \textcolor{gray}{63.86} & 98.38 / \textcolor{gray}{98.21} & 90.77 / \textcolor{gray}{90.21} & 99.16 / \textcolor{gray}{98.99} & 98.63 / \textcolor{gray}{98.43} & 82.43 / \textcolor{gray}{\textbf{81.81}} \\
    & Zeros & 9.30 / \textcolor{gray}{9.24} & 8.27 / \textcolor{gray}{9.09} & 72.43 / \textcolor{gray}{72.13} & 46.00 / \textcolor{gray}{43.27} & 96.44 / \textcolor{gray}{96.05} & 41.08 / \textcolor{gray}{40.49} & 45.59 / \textcolor{gray}{45.05} \\
    & Ones & 2.01 / \textcolor{gray}{1.95} & 1.16 / \textcolor{gray}{1.16} & 30.89 / \textcolor{gray}{26.26} & 10.29 / \textcolor{gray}{8.60} & 50.95 / \textcolor{gray}{46.61} & 3.73 / \textcolor{gray}{4.41} & 16.51 / \textcolor{gray}{14.83} \\
    \midrule
    & Regular LDU & 40.50 / \textcolor{gray}{40.44} & 65.12 / \textcolor{gray}{62.37} & 98.28 / \textcolor{gray}{98.20} & 90.61 / \textcolor{gray}{\textbf{90.61}} & 99.04 / \textcolor{gray}{98.95} & \textbf{98.92} / \textcolor{gray}{\textbf{98.92}} & 82.08 / \textcolor{gray}{81.58} \\
    \rowcolor{gray!20}
    & Uniform(\ding{62}) & 42.15 / \textcolor{gray}{39.72} & \textbf{66.66} / \textcolor{gray}{64.54} & \textbf{98.59} / \textcolor{gray}{98.28} & \textbf{91.21} / \textcolor{gray}{90.48} & \textbf{99.21} / \textcolor{gray}{98.97} & 98.63 / \textcolor{gray}{98.82} & \textbf{82.74} / \textcolor{gray}{81.80} \\
    \midrule
    \end{tabular}
    \end{adjustbox}
    \label{tab:init_ablation}
\end{table*}

\begin{table}[!h]
    \centering
    \footnotesize
    \setlength{\tabcolsep}{6pt}
    \caption{\textbf{Comparison of Image Generation Methods.} Performance metrics (DINO, CLIP-T, and CLIP-I) for DreamBooth, Textual Inversion, and LoLDU methods. Higher values indicate better performance. \textbf{Bold} values indicate best performance for each metric.}
    \begin{tabular}{@{}l|l@{}|rcccc@{}}
        \toprule
         Model & Method  & DINO $\uparrow$ & CLIP-T $\uparrow$ &  CLIP-I $\uparrow$  & Avg.\\ 
         \midrule 
         \multirow{4}{*}{\rotatebox{90}{SD-v1.4}} 
        & DreamBooth &  0.679 & \textbf{0.323} & 0.801 & 0.601 \\ 
        & Textual Inversion ~&  0.649 & 0.313 & 0.801 & 0.588 \\  
        \cmidrule(l){2-6}
        & \cellcolor{gray!20}\loldunospace ~& \cellcolor{gray!20}\textbf{0.723} & \cellcolor{gray!20}0.319 & \cellcolor{gray!20}\textbf{0.830} & \cellcolor{gray!20}\textbf{0.750} \\  
        \bottomrule
    \end{tabular}
    \label{tab:results-cv2}
  \end{table}

\paragraph{Results} We present the visual results in Figure \ref{fig:cv2},  while Table~\ref{tab:results-cv2} provides a quantitative comparison. 
We assess our method's efficacy through DINO, CLIP-T and CLIP-I metrics. 
DINO\cite{caron2021emergingpropertiesselfsupervisedvision} is computed as the average pairwise cosine similarity between the ViT-S/16 DINO embeddings of generated and real images. CLIP-I measures the average pairwise cosine similarity between CLIP\cite{radford2021learningtransferablevisualmodels} embeddings of generated and real images, while CLIP-T evaluates prompt fidelity by measuring the average cosine similarity between prompt and image CLIP embeddings. 
\loldu achieves the highest average score across metrics.

\begin{figure*}[t]
    \centering
    \vspace{-1em}
    \includegraphics[width=0.85\textwidth]{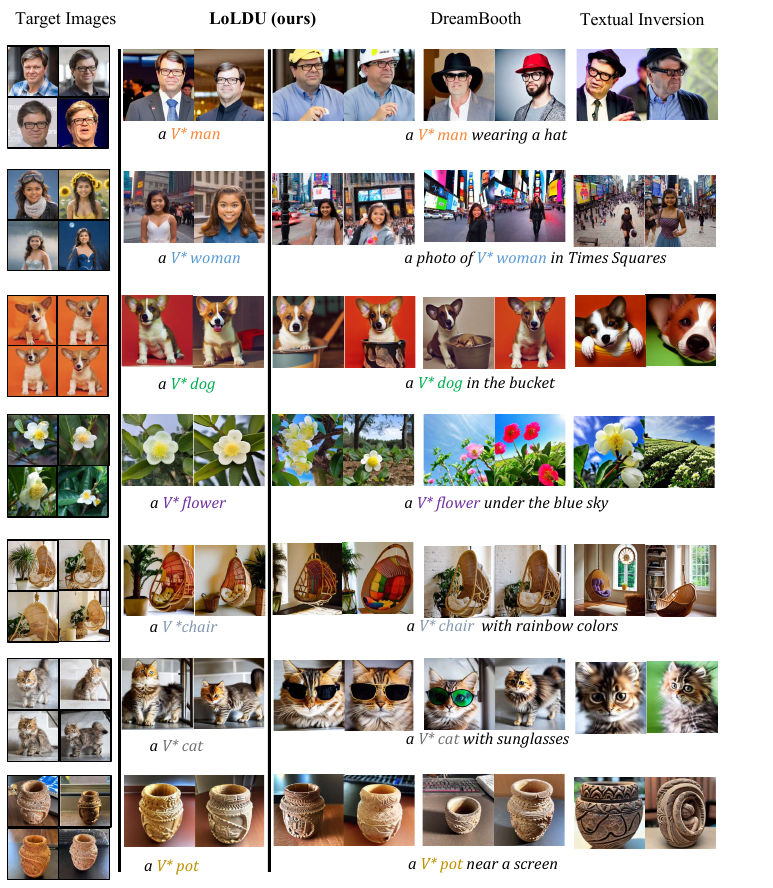}
    \caption{\textbf{Visualized Results of the Image Generation Task.}
    From left to right: target reference images, outputs from \loldu (ours), DreamBooth, and Textual Inversion. Each row represents a distinct category with a specified prompt (annotated under each row). \loldu demonstrates efficacy in generating diverse, prompt-adherent images while preserving key attributes from the reference set.}
    \label{fig:cv2}
\end{figure*}

\subsection{Analysis}
In this section, we conduct a comprehensive analysis of the hyperparameters associated with \loldunospace, specifically focusing on initialization, scaling factor, and rank. We systematically investigate the influence of these parameters on the performance and efficiency of our method across a variety of tasks.

\paragraph{Effect of Initialization}
The initialization of the entries $z$ in the diagonal matrix $diag(z)$ (Eq.~\ref{eq:1}) plays a crucial role in \loldunospace's performance. We evaluate several initialization policies on the ViT Base model across six image classification datasets. Table~\ref{tab:init_ablation} presents our findings.

Empirical results indicate that Uniform initialization consistently outperforms other strategies, achieving the highest average accuracy by stabilizing the training loop and enhancing convergence. Thus, \loldu with Uniform initialization is optimal for applications requiring stable dynamics and high accuracy. Additionally, both Uniform and Normal initialization contribute to training stability.
\paragraph{Impact of Scaling Factor}
The scaling factor within \loldu is crucial for assessing the efficacy of low-rank updates in augmenting model performance. This ablation study is dedicated to examining the necessity of integrating a scaling factor, specifically fixed at a value of 1, to evaluate its impact on enhancing model accuracy and ensuring training stability.

Table~\ref{tab:init_ablation} presents a comprehensive comparative analysis of performance metrics with and without the incorporation of a scaling factor across various datasets. The empirical findings reveal that the absence of a scaling factor, as denoted by the gray values, consistently leads to diminished accuracy and compromises the stability of the convergence process. This highlights the pivotal role of the scaling factor in optimizing the performance of \loldunospace, thereby enabling robust and efficient learning dynamics across a diverse range of image classification tasks.

\paragraph{Influence of Rank}
The rank parameter within \loldu is pivotal in determining the model's complexity and expressiveness. We conducted an extensive analysis by varying the rank across diverse tasks, as detailed in Table~\ref{tab:rank_ablation}. Additionally, the visual results of this analysis are presented in Figure~\ref{fig:rank}.

Our findings indicate that an increase in rank consistently enhances performance across all datasets, especially at lower ranks, but stabilizes beyond 256, indicating diminishing returns. Thus, selecting an optimal rank balances expressiveness and efficiency.
In practical applications of \loldunospace, our findings suggest that adopting a rank approximately one-third of the full rank ensures an optimal balance between performance and resource efficiency, thereby providing broader applicability across various scenarios.

\paragraph{Parameter Efficiency vs. Performance Trade-off}
Finally, we explore the nuanced relationship between parameter efficiency and performance, focusing on the capabilities of \loldu in comparison to other established methodologies. 

Table~\ref{tab:comparision} provides a compelling insight into the efficiency of \loldu, which achieves a mean accuracy of 82.79\% while utilizing a mere 0.21\% of the parameters. This is a stark contrast to methods like FullFT, which, despite achieving a higher accuracy of 88.20\%, require the full parameter set, and LoRA, which uses 6.77\% of the parameters for a lower accuracy of 76.22\%. These data underscore \loldunospace's exceptional capacity to deliver competitive performance with a substantially reduced parameter footprint.

\loldunospace's efficiency in parameter usage not only reduces computational and memory demands but also enhances the model's adaptability to various deployment scenarios, particularly those with limited resources. This efficiency is achieved without compromising on key performance metrics, as evidenced by the method's ability to maintain orthogonality, avoid random initialization, eliminate extra inference costs, and ensure faster convergence. These attributes collectively position \loldu as a highly effective and resource-efficient alternative to traditional methods, offering a strategic advantage in both research and practical applications.

\section{Conclusion}
\label{sec_5}
In conclusion, LoLDU represents a significant advancement in Parameter-Efficient Fine-Tuning (PEFT), offering a novel approach with the Lower-Diag-Upper (LDU) decomposition technique. By optimizing just 0.00025\% of parameters while maintaining performance across diverse tasks and model architectures, LoLDU addresses the prohibitive computational and storage costs associated with fine-tuning large models. Its preservation of orthogonality in triangular matrices and precise diagonal matrix optimization ensure efficient scale transformation and robust convergence. Our extensive evaluation, spanning various tasks and model scales up to 7 billion parameters, validates LoLDU's effectiveness and superiority over traditional fine-tuning methods, underscoring its potential for broad applicability and impact in advancing efficient model customization practices.

\bibliographystyle{IEEEtran}
\bibliography{IEEEabrv,references}

\clearpage
\appendix{\section{Appendix} \label{app}

This appendix provides supplementary material to support the methodologies and findings presented in the main manuscript. It is organized into five key areas: Natural Language Understanding, Instruction Tuning, Image Classification, Image Generation, and Ablation Studies. Each section offers detailed insights into datasets, experimental protocols, and hyperparameter settings, ensuring the replicability and validation of our results.

\begin{itemize}
    \item Section \ref{app:nlu}: Analysis of the GLUE benchmark and hyperparameters for Natural Language Understanding tasks.
    \item Section \ref{app:hp_llama}: Examination of the Alpaca dataset and LLaMA-2 model fine-tuning hyperparameters for Instruction Tuning.
    \item Section \ref{app:cv1}: Overview of image classification datasets and Vision Transformer (ViT) fine-tuning configurations.
    \item Section \ref{app:cv2}: Exploration of datasets for image generation and Stable Diffusion hyperparameters.
    \item Sections \ref{app:ablation}: Ablation studies on learning rate and rank variations affecting model performance.
\end{itemize}

\subsection{Natural Language Understanding} \label{app:nlu}
\subsubsection{GLUE Benchmark Details}
The GLUE benchmark is a framework for evaluating NLP models across nine tasks, such as CoLA, SST-2, and MRPC, focusing on grammaticality, sentiment, and semantic similarity. It includes a diagnostic dataset for assessing linguistic phenomena, aiding in the development of robust NLP systems through transfer learning. For more details, see the \href{https://medium.com/@researchgraph/introduction-to-glue-benchmark-82d1b7d161c8}{GLUE Benchmark Overview}.
\subsubsection{Hyperparameters for GLUE Experiments}
Table \ref{tab:hp_glue} details the hyperparameters for GLUE experiments.
\begin{table}[h!]
    \centering
    \small
    \setlength{\tabcolsep}{4pt}
    \renewcommand{\arraystretch}{1.1}
    \caption{Hyperparameters for GLUE Tasks}
    \label{tab:hp_glue}
    \begin{adjustbox}{max width=\columnwidth}
    \begin{tabular}{l|c|c|c}
        \toprule
        \textbf{Task} & \textbf{LR} & \textbf{Epochs} & \textbf{Max Length} \\
        \midrule
        MNLI & 3e-4 & 10 & 128 \\
        SST-2 & 4e-4 & 10 & 128 \\
        MRPC & 3e-4 & 20 & 512 \\
        CoLA & 2e-4 & 20 & 128 \\
        QNLI & 2e-4 & 10 & 512 \\
        QQP & 3e-4 & 20 & 512 \\
        RTE & 4e-4 & 20 & 512 \\
        STS-B & 2e-4 & 30 & 512 \\
        \midrule
        \multicolumn{4}{l}{Base: roberta-base, Batch: 32, Rank: 768, Alpha: 768} \\
        \multicolumn{4}{l}{Modules: query, value, Warmup: 0.06} \\
        \bottomrule
    \end{tabular}
    \end{adjustbox}
\end{table}

\subsection{Instruction Tuning} \label{app:hp_llama}
\subsubsection{Alpaca Dataset Overview}
The Alpaca dataset serves as a crucial asset for instruction tuning, consisting of 52,000 instruction-output pairs generated using OpenAI's `text-davinci-003` engine. Its primary goal is to improve the instruction-following capabilities of language models by providing a diverse array of instructional scenarios. The dataset is produced through the Self-Instruct framework, which includes modifications such as employing `text-davinci-003` for instruction generation and implementing aggressive batch decoding to enhance efficiency. The Alpaca dataset's diversity and high-quality annotations make it a valuable resource for training models to perform well across various tasks. This section explores the distinctive features of the Alpaca dataset, highlighting its role in the fine-tuning process of language models. For more details, refer to the \href{https://huggingface.co/datasets/tatsu-lab/alpaca}{Hugging Face dataset card for Alpaca}.
\subsubsection{Hyperparameters for LLaMA-2 Fine-tuning}
Table \ref{tab:hp_llama} provides a comprehensive overview of the hyperparameter settings employed during the fine-tuning of the LLaMA-2 model. These parameters are critical for optimizing model performance and ensuring robust convergence across various tasks.
\begin{table}[!h]
    \centering
    \small
    \setlength{\tabcolsep}{6pt}
    \renewcommand{\arraystretch}{0.90}
    \caption{Hyperparameters for Instruction Tuning}
    \begin{tabular}{l|c}
        \toprule
        \textbf{Hyperparameter} & \textbf{Value} \\
        \midrule
        Base Model & LLaMA2-7B \\
        Precision & BF16 \\
        Batch Size & 128 \\
        Micro Batch Size & 1 \\
        Learning Rate & 1e-3 \\
        Number of Epochs & 3 \\
        Rank & 1024 \\
        Alpha & 1024 \\
        Target Modules & q\_proj, v\_proj \\
        Cutoff Length & 256 \\
        Seed & 42 \\
        \bottomrule
    \end{tabular}
    \label{tab:hp_llama}
\end{table}

\subsection{Image Classification} \label{app:cv1}
\subsubsection{Dataset Descriptions}
This section introduces the datasets employed for image classification tasks, which include CIFAR10 \cite{cifar10}, EuroSAT \cite{eurosat}, StanfordCars \cite{cars}, FLOWERS102 \cite{flowers}, FGVC \cite{fgvc}, and CIFAR100 \cite{cifar100}. These datasets are selected to represent a broad spectrum of visual concepts and complexities, ranging from small to large label spaces.

\subsubsection{Hyperparameters for ViT Fine-tuning}
The hyperparameter settings utilized for the fine-tuning of the Vision Transformer (ViT) model are detailed in Table \ref{tab:hp_vit}.
\begin{table}[t]
    \centering
    \small
    \setlength{\tabcolsep}{6pt}
    \renewcommand{\arraystretch}{0.90}
    \caption{Hyperparameters for Image Classification}
    \begin{tabular}{l|c}
        \toprule
        \textbf{Hyperparameter} & \textbf{Value} \\
        \midrule
        Model & vit-b16-224-in21k \\
        Learning Rate & 3e-3 \\
        Batch Size & 128 \\
        Max Epochs & 10 \\
        Precision & bf16 \\
        Optimizer & AdamW \\
        LR Scheduler & Linear \\
        Warmup Steps & 30 \\
        Target Modules & query, value  \\
        Rank & 768 \\
        Alpha & 768 \\
        Seed & 42 \\
        \bottomrule
    \end{tabular}
    \label{tab:hp_vit}
\end{table}

\subsection{Image Generation} \label{app:cv2}
\subsubsection{Dataset Details}
The CustomConcept101 and Flickr-Faces-HQ (FFHQ) datasets provide concept images for fine tuning our image generation model. FFHQ contains 70,000 high-resolution images (1024×1024) with diverse attributes such as age, ethnicity, and accessories. Images were sourced from Flickr, aligned, and cropped using dlib, excluding non-human subjects. For more information, see the \href{https://github.com/NVlabs/ffhq-dataset}{FFHQ Dataset}.
\subsubsection{Hyperparameters for Stable Diffusion Fine-tuning}
The hyperparameter settings utilized for the fine-tuning of the Stable Diffusion model are detailed in Table \ref{tab:hp_sd}.
\begin{table}[htp]
    \centering
    \small
    \setlength{\tabcolsep}{6pt}
    \renewcommand{\arraystretch}{0.90}
    \caption{Hyperparameters for Image Generation}
    \begin{tabular}{l|c}
        \toprule
        \textbf{Hyperparameter} & \textbf{Value} \\
        \midrule
        Base Model & stable-diffusion-v1-5 \\
        VAE & sd-vae-ft-mse \\
        Learning Rate & 5e-4 \\
        Precision & fp16 \\
        Resolution & 512 \\
        Train Batch Size & 1 \\
        Optimizer & AdamW \\
        LR Scheduler & constant \\
        LR Warmup Steps & 15 \\
        Max Train Steps & 1000 \\
        Rank & 32 \\
        Alpha & 32 \\
        Seed & 42 \\
        Adam Weight Decay & 0.01 \\
        Target Modules & to\_k, to\_v, to\_q, to\_out \\
        \bottomrule
    \end{tabular}
    \label{tab:hp_sd}
\end{table}

\subsection{Ablation Studies}\label{app:ablation}
\subsubsection{Learning Rate} \label{app:lr}

\begin{figure}[!h]
    \centering
    \vspace{-1em}
    \begin{adjustbox}{max width=\textwidth} %
        \includegraphics[width=0.5\textwidth]{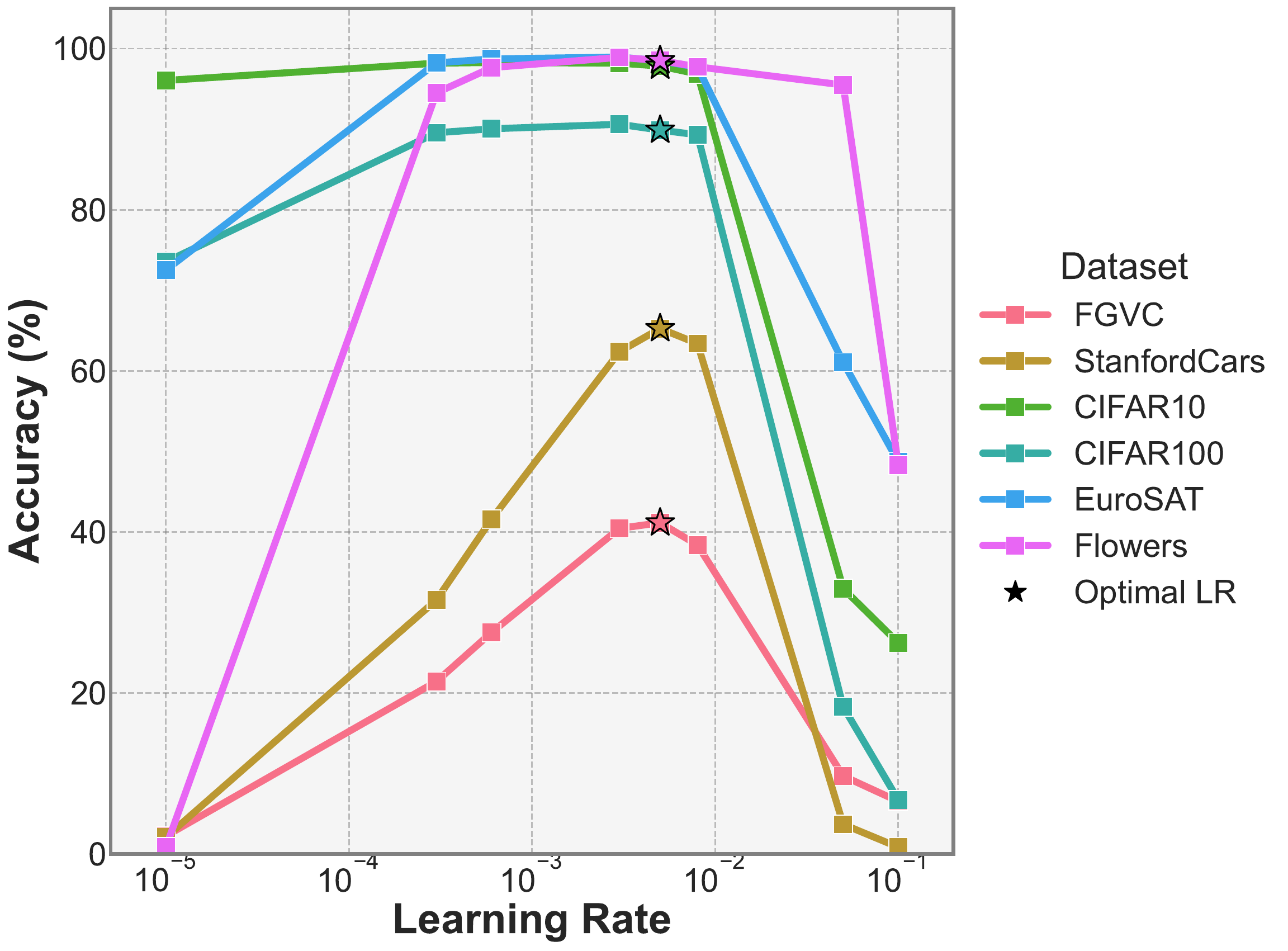}
    \end{adjustbox}
    \caption{\textbf{Learning Rate Ablation Study.} The figure demonstrates the effect of different learning rates on ViT-base model accuracy across FGVC, StanfordCars, CIFAR10, CIFAR100, EuroSAT, and Flowers datasets.}
    \label{fig:lr}
\end{figure}

\begin{table}[htp]
    \centering
    \scriptsize
    \setlength{\tabcolsep}{4pt}
    \caption{LR Ablation for ViT-Base: Comparison on FGVC, StanfordCars, CIFAR10, CIFAR100, EuroSAT, and Flowers. All ranks set to 768. \textbf{Bold} indicates best results.}
    \begin{tabular}{lccccccc}
    \toprule
    LR & FGVC & StanfordCars & CIFAR10 & CIFAR100 & EuroSAT & Flowers & Avg. \\
        & acc & acc & acc & acc & acc & acc & \\ 
    \midrule
    \multicolumn{8}{c}{\textbf{ViT-Base LR Ablation}} \\
    \midrule
    1e-1   & 6.54  & 0.85  & 26.21 & 6.71  & 48.70 & 48.31 & 22.89 \\
    5e-2   & 9.69  & 3.69  & 32.96 & 18.28 & 61.06 & 95.49 & 36.86 \\
    8e-3   & 38.37 & 63.38 & 96.86 & 89.30 & 97.69 & 97.75 & 80.56 \\
    \textbf{5e-3}   & \textbf{41.13} & \textbf{65.25} & 97.84 & 89.89 & 98.50 & 98.53 & \textbf{81.86} \\
    3e-3   & 40.44 & 62.37 & \textbf{98.20} & \textbf{90.61} & \textbf{98.95} & \textbf{98.92} & 81.58 \\
    6e-4   & 27.51 & 41.57 & 98.28 & 90.05 & 98.73 & 97.65 & 75.63 \\
    3e-4   & 21.42 & 31.55 & 98.20 & 89.56 & 98.23 & 94.51 & 72.25 \\
    1e-5   & 2.25  & 2.10  & 96.05 & 73.53 & 72.53 & 0.88  & 41.22 \\
    \bottomrule
    \end{tabular}
    \label{tab:lr_ablation}
\end{table}

This section provides an academic analysis of the impact of varying learning rates on model training. The visual representation, as detailed in \ref{fig:lr}, illustrates the outcomes of the learning rate ablation study, while the accompanying table, referenced in \ref{tab:lr_ablation}, provides comprehensive quantitative data.

\subsubsection{Rank Ablation} \label{app:rank}
\begin{table}[t]
    \centering
    \scriptsize
    \setlength{\tabcolsep}{4pt}
    \caption{ViT Rank Ablation Study on FGVC, StanfordCars, CIFAR10, CIFAR100, EuroSAT, and Flowers datasets. Different ranks indicate varying parameter counts. \#Params: Tunable parameters (M). The first section shows the base version, followed by the large-scale ablation. \textbf{Bold} denotes optimal LoLDU results.}
    \begin{tabular}{lc|cccccc}
    \toprule
    Rank & Params & FGVC & StanfordCars & CIFAR10 & CIFAR100 & EuroSAT & Flowers \\
    \midrule
    \multicolumn{7}{c}{\textbf{ViT-Base Rank Ablation}} \\ 
    \midrule
    1    & 24     & 27.59 & 43.95 & 96.81 & 86.67 & 95.25 & 98.33 \\
    8    & 192    & 28.28 & 48.40 & 97.47 & 89.84 & 96.14 & 98.53 \\
    16   & 384    & 31.13 & 50.87 & 97.76 & 88.48 & 96.74 & 98.53 \\
    32   & 768    & 32.75 & 53.00 & 97.82 & 88.76 & 97.28 & 98.63 \\
    64   & 1536   & 34.01 & 55.09 & 97.96 & 89.60 & 97.60 & 98.53 \\
    128  & 3072   & 34.91 & 58.20 & 98.07 & 89.89 & 98.20 & 98.53 \\
    256  & 6144   & 36.38 & 61.44 & 98.06 & 90.18 & 98.62 & 98.63 \\
    512  & 12288  & 38.48 & 63.68 & 98.17 & 90.30 & 98.83 & 98.63 \\
    \rowcolor{gray!20}
    \textbf{768} & 18456 & \textbf{42.15} & \textbf{66.66} & \textbf{98.59} & \textbf{91.21} & \textbf{99.21} & \textbf{98.63} \\
    \midrule
    FT   & 85.8  & 54.84 & 79.78 & 98.92 & 92.38 & 99.05 & 98.43 \\
    LoRA & 581   & 25.16 & 45.38 & 98.78 & 92.02 & 98.44 & 97.55 \\
    \bottomrule
    \end{tabular}
    \label{tab:rank_ablation}
\end{table}

This subsection presents an analysis of the rank ablation study, examining the impact of different parameter ranks on model performance. Table \ref{tab:rank_ablation} summarizes the results.

}

\vfill

\end{document}